\pdfoutput=1

\documentclass[11pt]{article}

\usepackage[]{acl}
\usepackage{times}
\usepackage{latexsym}
\usepackage{graphicx}
\usepackage[T1]{fontenc}

\usepackage[utf8]{inputenc}

\usepackage{microtype}

\title{BFCAI at SemEval-2022 Task 6: Multi-Layer Perceptron for Sarcasm Detection in Arabic Texts}

 \author{Nsrin Ashraf \and Fathy Elkazaz \and Mohamed Taha \and  Hamada Nayel \\
       Department of Computer Science, Benha University, Benha, Egypt\\
       \texttt{\{nisrien.ashraf19,fathy.elkazzaz\}@fci.bu.edu.eg}\\
       \texttt{\{mohamed.taha, hamada.ali\}@fci.bu.edu.eg}\\
      { \bf Tarek Elshishtawy}\\
        Department of Information System, Benha University, Benha, Egypt\\
        \texttt{t.shishtawy@fci.bu.edu.eg}}

\begin{document}
\maketitle
\begin{abstract}
This paper describes the systems submitted to iSarcasm shared task. The aim of iSarcasm is to identify the sarcastic contents in Arabic and English text. Our team participated in iSarcasm for the Arabic language. A multi-Layer machine learning based model has been submitted for Arabic sarcasm detection. In this model, a vector space TF-IDF has been used as for feature representation. The submitted system is simple and does not need any external resources. The test results show encouraging results.
\end{abstract}
\section{Introduction}
Analyzing social media becomes a crucial task, due to the frequently usage of social media platforms. Sarcasm detection, the conflict of using the verbal meaning of a sentence and its intended meaning \cite{clift_1999,doi:10.1207/s15327868ms1301}, is an important task. Sarcasm detection is a challenge, since sarcastic contents are used to express the opposing of what is being said. Recently sarcasm detection has been studied from a computational perspective as one of classification problems that separates sarcastic from non-sarcastic contents\cite{Reyes:2013:MAD:2447287.2447294,nayel-etal-2021-machine-learning}. 
\par{}Arabic is an important natural language having an extensive number of speakers. The research in Natural Language Processing (NLP) for Arabic is continually increasing. However, there is still a need to handle the complexity of NLP tasks in Arabic. This complexity arises from various aspects, such as orthography, morphology, dialects, short vowels, and word order. Sarcasm detection in Arabic is a particularly challenging task \cite{Alayba_2018}. 
\par{}In this paper, we describe the system submitted to the iSarcasm detection shared task\cite{abufarha-etal-2022-semeval}. The shared task aims at detecting the sarcasm contents in Arabic tweets. In this work, a machine learning framework has been developed and various machine learning algorithms have been implemented. Term Frequency-Inverse Document Frequency (TF-IDF) has been used as vector space model for tweet representation. The rest of this paper is organized as follows: in section 2, a background about sarcasm detection is given. Section 3 and section 4 overview the dataset and the system respectively. Experimental setup and results are given in section 5 and section 6 respectively. Finally, section 7 concludes the proposed work and suggests future work to be continued. 
\section{Background}
The research work have been done on Arabic sarcasm detection were mainly focused on creating datasets and establish a baseline for each created dataset \cite{10.1007/978-3-030-45442-5_18}. \citet{karoui2017soukhria} created a corpus of sarcastic Arabic tweets that are related to politics. Distant supervision has been used for the creation of corpus. The authors used keywords that are like sarcastic contents in Arabic to label the tweets as sarcastic tweets. They implemented different machine learning algorithms such as SVM, logistic regression, Naïve Bayes, and other classifiers on the developed corpus.
\par{}An ensemble classifier of XGBoost, random forest and fully connected neural networks has been designed by \citet{khalifa-2019-ensemble}. They extracted a set of features that consists of sentiment and statistical features, in addition to word $n$-grams, topic modelling features and word embeddings. \citet{nayel-2019-benha} developed an ensemble-based system for irony detection in Arabic Tweets. A set of classification algorithms namely Random forests, multinomial Naïve Bayes, linear, and SVM classifiers have been used as base-classifiers. In \cite{nayel-etal-2021-machine-learning}, sarcasm detection has been formulated as a binary classification problem and SVM has been implemented. 
\section{Dataset}
A new data collection method has been introduced, where the sarcasm labels for texts are provided by the authors themselves. The author of each sarcastic text rephrased the text to convey the same intended message without using sarcasm. \citet{leggitt2000emotional} defined a set categories of ironic speech namely; sarcasm, irony, satire, understatement, overstatement, and rhetorical question. Linguistic experts have been asked to further label each text into one of these categories. Each text in the Arabic dataset has the following information attached to it:
\begin{itemize}
\item a label specifying the text dialect;
\item a label specifying the nature of sarcasm (sarcastic or non-sarcastic), provided by its author;
\item a rephrase provided by its author that conveys the same message non-sarcastically.
\end{itemize}
\section{System Overview}
\begin{figure}[htb!]
\centering\includegraphics[height=2.8in,width=0.5\textwidth]{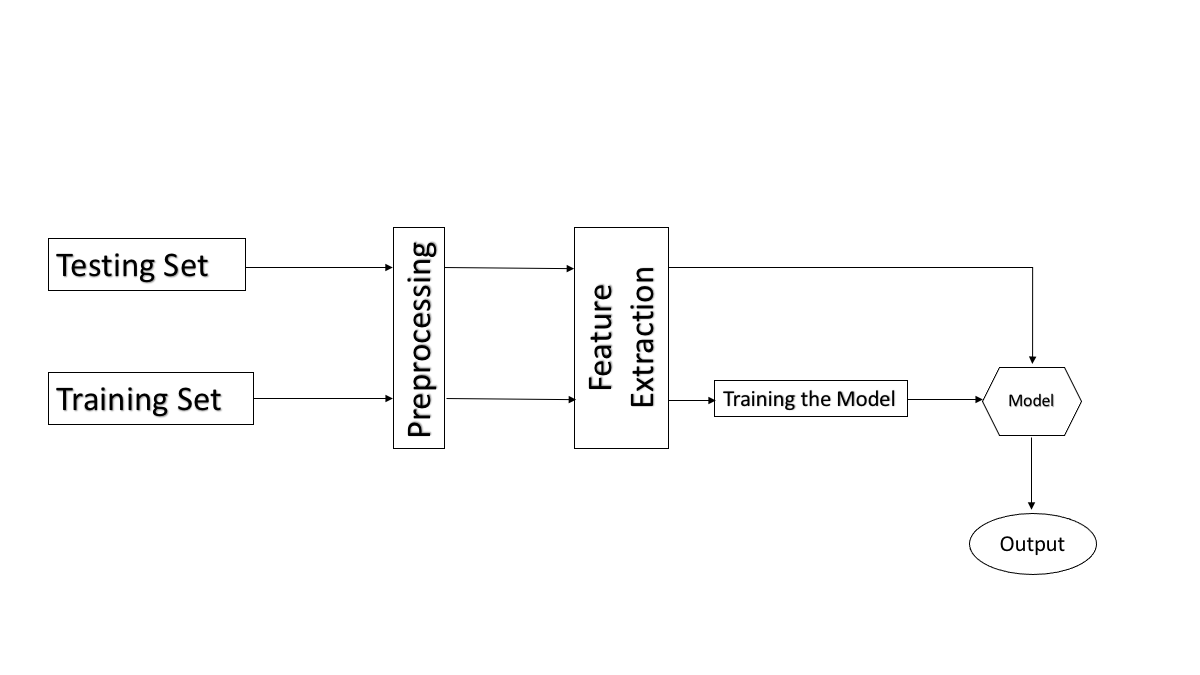}
\caption{\label{diagram}The structure of the proposed model}
\end{figure}

In this section, we review the main structure of the proposed model. The proposed system, as shown in figure \ref{diagram}, consists of three phases namely; preprocessing, feature extraction and training the classification algorithms. Then, the resulted model used to predict the unseen test data. 
\subsection{Preprocessing}
The first stage of developing systems is preprocessing, where unwanted and uninformative piece of text has to be removed, it is also called text cleaning. We performed text cleaning by removing:
\begin{itemize}
\item special symbols, such as $\{+, -, =, \$,....\}$;
\item repeated characters such as ("\emph{hhhhhhhh}" will be normalized to "\emph{hh}"); 
\item non-Arabic words, such as English characters or any other language;
\item punctuations and Arabic diacritics.
\end{itemize}
\subsection{Features Extraction}
To prepare features to build classification model and before feeding the text into the classifier and after performing text cleaning,  Term Frequency-Inverse Document Frequency (TF-IDF) technique was used to change over content to vectors and all the algorithms to investigate the best performing algorithm.\\
\indent TF-IDF has been used to represent comments as vectors. If $ <\!\!\!w_1,w_2, \ldots, w_k\!\!\!> $ are the tokenized words of a comment $ \mathcal{T}_j$, the vector associated to the comment $ \mathcal{T}_j$ will be represented as $<\!\!\!v_{j1}, v_{j2},\ldots ,v_{jk}\!\!\!>$ where $v_{ji}$ is the weight of the token $w_i$ in tweet $ \mathcal{T}_j$ which is calculated as:- \[v_{ji} = tf_{ji} * \log\left(\frac{N+1}{df_i+1}\right)\] where $tf_{ji}$ is the total number of occurrences of token $w_i$ in the comment $ \mathcal{T}_j$, $df_i$ is the number of comments in which the token $w_i$ occurs and $N$ is the total number of comment.\\
\subsection{Methodology} 
We explored various classification algorithms as well as ensemble approach by combining the output of these classifiers (also known as base classifiers) using hard voting. The base classifiers used in this work are listed below:
\begin{itemize}
\item{Support Vector Machines (SVMs)} 
\item{Random Forest (RF)}
\item{K-Nearest Neighbours (KNN)}
\item{Multinomial Naïve Bayes (M-NB)}
\item{Multi-Layer Perceptron (MLP)}
\item{Stochastic Gradient Descent (SGD)}
\item{AdaBoost Classifier}
\item{Voting Classifier}
\end{itemize}
\section{Experimental Setup}
For feature extraction phase we used unigram model. For the purpose of training the model, we have used 5-fold cross-validation technique to adjust the parameters.\\
\indent The Scikit-Learn library implementation of classification algorithms were used in the training phase. \indent For SVM, two kernels have been tested: linear kernel and RBF with two parameters $\gamma = 2 $ and $C = 1$. While, for SGD classifier the loss function used was Hinge and the maximum iteration was set at 10000 iterations.\\
\indent The number of nodes in the hidden layer of MLP was set at 20, logistic function was used as activation function and Adam solver was used. The maximum number of decision trees in random forests is set at 300.
\subsection{Evaluation Metrics}
\par{}F1-score has been used to evaluate the performance of all submissions. F1-score is a harmonic mean of Precision (P) and Recall (R) and calculated as follow: 
\[ F\!\!-\!\!score = \frac{2*P*R}{P + R }\]
F1-score for the sarcastic class (F1-sarcastic) has been used for final evaluation.
\section{Results}
The cross validation accuracy of all training classifiers for the training set is given in Table \ref{tab1}. It is clear that MLP gives the best accuracy with moderate Standard Deviation (STD) for the five folds while development phase, so we decided to submit the output of this classifier.\\
\begin{table}
\centering
\caption{5-fold Cross-Validation accuracy for all classifiers in the training set}\label{tab1}
\begin{tabular}{l|c|c}\hline
Classifier & Accuracy &  STD \\\hline
 SVM-Linear&81.0\% & 0.055  \\
 SVM-RBF&76.6\% & 0.003  \\
 MNB&76.6\% & 0.005\\
 SGD&80.0\% & 0.045\\
 MLP&83.6\% & 0.045\\
 RF&75.8\% & 0.056\\ 
 KNN &79.7\% & 0.058\\
AdaBoost &75.2\% & 0.052\\
Voting&80.4\% & 0.043\\\hline
\end{tabular}
\end{table}

\begin{table}
\centering
\caption{Results of MLP classifier for the test set }\label{tab2}
\begin{tabular}{l|c|c}\hline
Measure & Value &  Rank \\\hline
F-1 sarcastic & 0.3746 & 14  \\
F-score & 0.6024 & 11  \\
Precision (P) & 0.5968 &15\\
Recall (R) &0.6608 & 17\\
Accuracy&0.7329 & 8\\\hline
\end{tabular}
\end{table}
\indent Results for test set is given in Table \ref{tab2}. The reported results show that, while training MLP gives better accuracy among implemented machine learning classifiers. Also, it gives better rank in accuracy for the unlabelled test set. While in other metrics, the performance was not satisfied. This may resulted because of using accuracy metric while comparing different classifiers in development phase.\\
\indent A good suggestion is to use different evaluation metrics while developing the system. In addition, using different word representation models such as word embeddings, which encompasses the semantic meaning of words could improve the performance. Complex models such as deep learning models is promising, but the challenge in such models is the availability of suitable resources.    
\section{Conclusion}
In this work, a classical machine learning framework has been designed for sarcasm detection in Arabic tweets. The proposed framework reported reasonable results. The future work may include applying complex framework such as deep learning structure. In addition, word representation is very important factor that can be used in different manner such as word embeddings and transformers. 
\bibliography{ref}
\end{document}